\documentclass[twocolumn,10pt]{asme2e}

\confshortname{DSCC2020}
\conffullname{the ASME 2020 \&\\
              Dynamic Systems and Control Conference}

\confdate{4-7}
\confmonth{October}
\confyear{2020}
\confcity{Pittsburgh, PA}
\confcountry{USA}

\papernum{DSCC2020-3208}

\title{Contact-Rich Trajectory Generation in Confined Environments Using Iterative Convex Optimization}
\usepackage{balance}  
\usepackage{algorithm}
\usepackage{bm}
\usepackage{mathrsfs}
\usepackage{algorithm}
\usepackage{algpseudocode}
\usepackage{amsmath}
\usepackage{bm}
\usepackage{amsfonts}
\usepackage{cleveref}
\usepackage{graphicx}
\usepackage{framed}
\usepackage{subcaption}
\usepackage{array}
\usepackage{booktabs}
\usepackage{eucal}
\usepackage{flushend}

\usepackage{comment}
\usepackage{array}

\newcolumntype{L}[1]{>{\raggedright\arraybackslash}p{#1}}
\algnewcommand{\algorithmicor}{\textbf{ or }}
\algnewcommand{\OR}{\algorithmicor}
\newcommand{\etal}{\textit{et al}.}

\author{Wei-Ye Zhao
    \affiliation{
	Robotics Institute\\
	Carnegie Mellon University \\
	Pittsburgh, PA, 15213\\
    Email: weiyezha@andrew.cmu.edu
    }	
}

\author{Suqin He
    \affiliation{
	Robotics Institute\\
	Carnegie Mellon University \\
	Pittsburgh, PA, 15213\\
    Email: suqinh@andrew.cmu.edu
    }	
}

\author{Chengtao Wen
    \affiliation{
	Siemens CT\\
	Berkeley, CA, 94704\\
    Email: chengtao.wen@siemens.com
    }	
}

\author{Changliu Liu\thanks{This work was supported in part by Subaward No. ARM-TEC-18-01-F-06 from the Advanced Robotics for Manufacturing ("ARM") Institute under Agreement Number W911NF-17-3-0004 sponsored by the Office of the Secretary of Defense. ARM Project Management was provided by Matthew S. Fischer. The views and conclusions contained in this document are those of the authors and should not be interpreted as representing the official policies, either expressed or implied, of either ARM or the Office of the Secretary of Defense of the U.S. Government. The U.S. Government is authorized to reproduce and distribute reprints for Government purposes, notwithstanding any copyright notation herein. This project is in collaboration with Siemens and Yaskawa.}
    \affiliation{
	Robotics Institute\\
	Carnegie Mellon University \\
	Pittsburgh, PA, 15213\\
    Email: cliu6@andrew.cmu.edu
    }	
}

\begin{document}

\maketitle    

\begin{abstract}
Applying intelligent robot arms in dynamic uncertain environments (i.e., flexible production lines) remains challenging, which requires efficient algorithms for real time trajectory generation. The motion planning problem for robot trajectory generation is highly nonlinear and nonconvex, which usually comes with collision avoidance constraints, robot kinematics and dynamics constraints, and task constraints (e.g., following a Cartesian trajectory defined on a surface and maintain the contact). The nonlinear and nonconvex planning problem is computationally expensive to solve, which limits the application of robot arms in the real world. In this paper, for redundant robot arm planning problems with complex constraints, we  present a motion planning method using iterative convex optimization that can efficiently handle the constraints and generate optimal trajectories in real time. The proposed planner guarantees the satisfaction of the contact-rich task constraints and avoids collision in confined environments. Extensive experiments on trajectory generation for weld grinding are performed to demonstrate the effectiveness of the proposed method and its applicability in advanced robotic manufacturing.
\end{abstract}

\section*{INTRODUCTION}
In recent years, intelligent robot arms have been playing increasingly important roles in both industry and people's daily life. Nonetheless, one of the biggest challenges toward the wide adoption of intelligent robot arms still lies in real time motion planning,  i.e., how we can enable a robot to compute its trajectories to achieve its goal in real time~\cite{latombe2012robot}. Regardless of different configurations and sizes of the robots, to work in complex environments, they need to satisfy multiple task constraints such as collision avoidance and contact maintenance, which make the motion planning problem difficult to solve in real time.

Task constrained motion planning for robot arms rises in many situations~\cite{stilman2007task}, for example, when robots are interacting with humans, or when robots are interacting with the environment during door-opening, welding, polishing or grinding. The task constraints can  be divided into two categories: 1) the workspace inequality constraints (e.g., for collision avoidance), and 2) the task equality constraints (e.g., for contact maintenance). The workspace inequality constraints require that the robot does not collide with itself or with the  obstacles in the environment~\cite{mi2019sampling}. The task equality constraints can be interpreted as additional objectives that the robot should satisfy. For example, in a grinding task, the robot needs to maintain contact with the surface being grind. 

\subsection*{Related Work}
Conventional task-constrained motion planning uses kinematic control techniques~\cite{klein1983review} by transforming a task-space trajectory (or a Cartesian space trajectory) to the robot configuration space using the inverse of the task Jacobian. Since the configuration space usually has higher dimension than the task space, there is a null space associated with the Jacobian inverse. The motions in the null space can be optimized locally to achieve other objectives, i.e., collision avoidance. There are many methods for local null space optimization. The most widely used approaches are through energy function based methods~\cite{wei2019safe}, which include potential field methods~\cite{khatib1986real}, control barrier functions~\cite{Chiara2019ECC}, safe sets~\cite{HsienChung2017CCTA}, etc. Those methods first define a scalar energy function (also called a potential function, a barrier function, or a safety index) that attains small value in the collision free space and high value in the unsafe space. Then the scalar energy function will serve as the objective function to be minimized for the null space optimization. The resulting robot trajectory will be repulsed from the obstacles.  However, it is challenging and time consuming to generate the scalar energy function in the high-dimensional joint configuration spaces~\cite{bansal2017hamilton}. Moreover, the solution of the local optimization can easily be trapped into local optima, hence creating stability issues or deadlocks. 

To address the issues mentioned above, global planning approaches are needed. There are two types of algorithms  for global constrained motion planning: sampling-based methods~\cite{kim2016tangent} and optimization-based methods~\cite{liu2017convex}. Sampling-based methods plan trajectories by generating random joint space displacements until the goal is reached. Representative methods include probability road maps (PRM)~\cite{kavraki1996probabilistic} and rapidly-exploring random tree (RRT)~\cite{lavalle1998rapidly}, both of which can generate trajectories that satisfy the workspace inequality constraints (e.g., collision avoidance). However, the trajectories planned by construction are usually not smooth~\cite{li2019dynamical}. The chance for the random sampled trajectories to satisfy the task equality constraints (e.g., for contact maintenance) is low~\cite{stilman2007task}. Hence, the exploration may be inefficient.

Optimization-based methods, on the other hand, generate much smoother trajectories compared to the sampling-based methods. For optimization-based planning, an optimization problem needs to be formulated first, which includes an object function and multiple constraints. A desired trajectory is obtained by solving the optimization problem~\cite{liu2018convex}. For example, the CHOMP algorithm~\cite{ratliff2009chomp} solves motion planning problems that penalize the magnitude of joint velocities and accelerations and the distance towards static obstacles. The ITOMP algorithm~\cite{park2012itomp} divides obstacles into dynamic obstacles and static obstacles, and solves motion planning problems that penalize the distance towards both dynamic and static obstacles.

One major challenge for optimization-based planning algorithm is its computational inefficiency. Since the optimization problem for motion planning is usually highly nonlinear and non-convex, it is computationally expensive to obtain a solution using generic nonlinear optimization solvers such as sequential quadratic programming (SQP)~\cite{boggs1995sequential} and sequential expanded Lagrangian homotopy (SELH) ~\cite{Dharmawan2018JMR}. Those algorithms obtain solutions by solving the Karush-Kuhn-Tucker (KKT) equations. These generic algorithms work poorly on robot motion planning problems since they fail to incorporate domain specific information, e.g., the geometry of the problem. 

To improve the performance of optimization solvers and tackle the computation challenge, various methods that directly convexify the optimization problem using domain knowledge have been proposed. For example, Liu \etal~\cite{liu2018convex} proposed the convex feasible set (CFS) algorithm to efficiently handle non-convex inequality constraints through iterative convexification of the constraints. E. Todorov \etal~\cite{Emanuel2004ICINCO} proposed the iterative LQR algorithm to efficiently handle nonlinear equality constraints through iterative linearization of the constraints. Howell \etal~\cite{howell2019altro}~\cite{jackson2020scalable} proposed ALTRO which combines iLQR with an augmented Lagrangian method to handle general state and input constraints. However, the task equality constraints are not considered in ALTRO. To the best knowledge of the authors, there is very limited research on real time optimization algorithms for constrained trajectory generation that can handle both the non-convex workspace inequality constraints and the nonlinear task equality constraints. 

\subsection*{Challenges and Contributions}
This paper focuses on developing optimization-based planning algorithms for contact-rich trajectory generation in confined environments, where the robot trajectory in the joint space is subject to workspace inequality constraints for collision avoidance and equality constraints to maintain contact. This algorithm is applied on a weld bead removal task using a six degrees of freedom industrial robot arm. In this application, the workspace inequality constraints require the trajectory to be collision-free and the task equality constraints require the robot end-effector tip to always maintain contact with the weld bead. In addition to weld bead removel, the method for contact-rich trajectory generation in confined environments can also be applied to many other applications, such as robotic welding, rehabilitation, etc. 

There are two major challenges regarding the development of optimization-based contact-rich trajectory generation algorithms: 1) real time computation, 2) satisfaction of various constraints. As pointed out earlier, the optimization problem for trajectory generation in a cluttered environment is highly nonlinear and non-convex, which is hard to solve in real time. 

To address the aforementioned challenges, we propose an iterative convex optimization approach to efficiently handle the constraints and achieve real time computation.
To handle the constraints, we first leverage the CFS algorithm to iteratively 
transform the non-convex inequality constraints into a sequence of convex inequality constraints called the convex feasible sets~\cite{liu2017convex}. Then we leverage the approach in iLQR to iteratively linearize the nonlinear equality constraint. Hence the original nonlinear and non-convex optimization problem can be transformed into convex problems and solved iteratively. In the following discussion, we will call the proposed approach as iterative convex optimization for planning (ICOP).
Experimental results show that ICOP can generate desired trajectories that satisfies all the constraints in real time. 

The contributions of this paper are summarized as follows:
\begin{enumerate}
    \item We propose the ICOP framework, which can generate high quality trajectories with significantly reduced computation time compared to conventional methods. 
    \item We conduct real world robot polishing experiments to demonstrate the applicability of our proposed method in advanced robotic manufacturing.
\end{enumerate}

The remainder of the paper is organized as follows: In Section II, the problem for contact-rich trajectory generation in confined environments is formulated; in Section III, the proposed approach will be introduced; in Section IV, the performance of the proposed algorithm will be illustrated under a weld bead removal task and will be compared against conventional methods; Section V concludes the paper.

\section*{PROBLEM FORMULATION}
This paper  focuses on contact-rich trajectory generation in confined environment. The robot state is denoted as $x \in X \subset \mathbb{R}^n$, where $X$ is the configuration (state) space and  $n$ is its dimension, which specifies the degree of freedom of the robot. The robot state at a discrete time step $t$ is denoted as $x_t$. A trajectory is defined to be the sequence of states from time 1 to time T: $\bold{x} = [x_1;x_2;...;x_T] \in \mathbb R^{n \times T}$.

\subsection*{Safety Specification in Confined Environments}
The workspace inequality constraints are defined so that the robot should be collision-free with the surrounding obstacles. Suppose the area occupied by the obstacle in the Cartesian space is $O\subset\mathbb{R}^3$, the safety constraint for robot with state $x_t$ at time step $t$ is defined as: 
\begin{equation}\label{eq: safety}
d(x_t, O) > 0,
\end{equation}
where the function $d(\bullet): \mathbb{R}^n \times \mathbb{R}^3 \to \mathbb{R} $ is a signed distance function that computes the distance from the robot to the obstacle in the Cartesian space. To ensure safety, we require the distance to be greater than zero.

\begin{algorithm*}[h]
  \caption{Iterative Convex Optimization for Planning}\label{ICOP}
  \begin{algorithmic}[1]
    \Procedure{Iterative Convex Optimization for Planning}{$\mathbb{C}_{target},O,T,x_{ini},\xi$}
      \State \textbf{Inputs:}
      \State $\mathbb{C}_{target} \in \mathbb{R}^{m\times T}$: Pre-defined $j$-th robot body point trajectory in Cartesian space
      \State $O$: Environment obstacles
      \State $T$: planning horizon
      \State $x_{ini}$: initial robot joint state
      \State $\xi$: equality threshold 
      
      \State \textbf{Initialization:}
      \State $x_{pre} = x_{ini}$
	
	\State \textbf{Iteration:}
      \For{$t=0,1,2,..., T$} \Comment{Outer Loop}
      	\State Set $C_{next}\leftarrow\mathbb{C}_{target}(t)$, $x_{ref} \leftarrow x_{pre}$, and $C_{ref} \leftarrow \Gamma(x_{pre})$
	\While{$\| C_{next} - C_{ref} \| > \xi$ \OR $d(x_{ref}, O) < 0$} \Comment{SafeTrack Inner Loop}
		
		\State Find convex feasible set using \eqref{eq8}
		\State Linearize the nonlinear equality constraint using \eqref{eq9} 
		\State Solve the convex optimization problem from \eqref{eq10}, whose solution is ${\hat{x}}_{ref}$;
		\State Update $x_{ref} \leftarrow {\hat{x}}_{ref}$ and $C_{ref}\leftarrow \Gamma(\hat{x}_{ref})$
	\EndWhile 
	\State Record $\bm{x}(t) \leftarrow x_{ref}$ and $x_{pre} \leftarrow x_{ref}$
      \EndFor  
      \State \textbf{return} reference trajectory $\bm{x}$
    \EndProcedure
  \end{algorithmic}
  \label{alg1}
\end{algorithm*}

\subsection*{Contact-Rich Task Specification}
The contact-rich task equality constraints are defined so that the robot body is in contact with the specified targets. Mathematically, the constraint can be written as:
\begin{equation}
\Gamma_j(x_t) = p_j^t,
\label{2}
\end{equation}
where the $\Gamma_j(\bullet): \mathbb{R}^n \to \mathbb{R}^m $ is a generalized function to project the $j$-th point on the robot body (e.g., tool tip) to the constrained task space given the robot configuration $x_t$. $m\in\{1,2,\ldots, 6\}$ is the dimension of the task space. $m=2$ means the task is constrained in a 2-dimensional plane, e.g., wiping a surface. $m=3$ means the task constraints are defined in a 3-dimensional Cartesian space, e.g., welding on a predefined trajectory. And $m=6$ means the task has 3 translation constraints and 3 angular constraints. $p_j^t \in \mathbb{R}^m$  is predefined target for the $j$-th point on the robot body at time step $t$. Note that the mapping $\Gamma_j$ involves highly nonlinear forward kinematics. This paper mainly considers the case $m=3$.

\subsection*{Optimization Objective}
Instead of optimizing the trajectory as a whole, we propose to generate the contact-rich reference trajectory step by step. To avoid the potential instability and deadlocks issues for local optimization, we may perform the step-by-step optimization from time 1 to time T multiple rounds. 
The optimization problem at time step $t$ is formulated as: 
\begin{equation}
\begin{aligned}
& \underset{x}{\text{min}}
& & J(x_t,x_{t-1}) = \| x_t - x_{t-1}\|_Q^2 \\
& \text{s.t.} & &  \Gamma_j(x_t) = p_j^t, \forall j = 1,2,3,...,\\
& & &  d(x_t, O) > 0,\\
& & &  x_{min} \leq x_t \leq x_{max},\\
\end{aligned}
\label{eq:opt}
\end{equation}
where $J$ is the cost function; $\| x_t - x_{t-1}\|_Q^2 = (x_t - x_{t-1})^T Q (x_t - x_{t-1})$ penalizes the differences of states between two time steps; $Q$ is the weight matrix; $x_{t-1}$ is the reference configuration state from the last time step; $x_{min},x_{max}\in\mathbb{R}^n$ are joint limits.  The intuition behind \eqref{eq:opt} is that we want to minimize the norm difference between the joint position in the last step and the joint position in the current step to ensure that the robot joint states evolve smoothly. The three constraints are: 1) the equality constraints to maintain contact, 2) the inequality constraints to ensure safety, and 3) the robot joint limits.

\section*{METHODOLOGY}

This section discusses the methodology to solve \eqref{eq:opt}. We will first introduce iterative linearization of the nonlinear equality constraints, then discusse the convex feasible set algorithm to handle the nonconvex inequality constraints, and finally introduce the proposed  ICOP framework.

\subsection*{Iterative Equality Linearization Approximation}
The equality constraint in \eqref{2} encodes the task requirements, but is highly nonlinear and expensive to resolve. In this paper, we assume $\Gamma$ is a twice continuously differentiable function whose second derivative exists and is continuous. 
To speed up the computation, we propose to iteratively consider the first order approximation of the nonlinear equality constraint until the solution converges.

Suppose the initial joint state is $x_0$, the corresponding $j$-th robot body point is located at $C_0$. Our target is to find $x_1$ that satisfies $\Gamma_j(x_1) = C_1$, where $C_1$ is the next desired $j$-th robot body point. Suppose the distance between $C_1$ and $C_0$ is less than a small positive constant. Then we can use first-order linear approximation to represent $C_1$ as:
\begin{align}
C_1 &= \Gamma_j(x_0) + \nabla\Gamma_j(x_0)\cdot(x_1 - x_0) + \sigma, \\
	&= C_0 + \nabla\Gamma_j(x_0)\cdot(x_1 - x_0) + \sigma, \label{eq5}
\end{align}
where $\sigma$ is an error term, and $\sigma \to 0$ as $\| x_1 - x_0 \|_2^2 \to 0 $.  Denote $\nabla \Gamma_j(x)$ as $Jac(x, \theta)$, which is the generalized Jacobian matrix at $x$ with respect to a robot feature vector $\theta$ such as the DH parameters. Now we rewrite \eqref{eq5} as:
\begin{equation}
Jac(x_0,\theta)\cdot x_1 = Jac(x_0,\theta)\cdot x_0 + C_1 - C_0 - \sigma,
\label{iela}
\end{equation}
which is a linearized equality constraint for \eqref{2}. Inspired by iLQR, it is reasonable to assume that solving the optimization \eqref{eq:opt} with respect to the iterative approximation of nonlinear equality constraints in \eqref{iela} will lead to converging results~\cite{Emanuel2004ICINCO}. The formal convergence proof is beyond the scope of this paper, which is left for future work.

\subsection*{Convex Feasible Set Algorithm}
To deal with the nonlinear inequality constraint \eqref{eq: safety} for safety in confined environments, we leverage the Convex Feasible Set Algorithm (CFS)~\cite{liu2018convex} to efficiently search the non-convex feasible space for solutions by solving a sequence of convex optimizations constrained in the convex feasible sets. 

The CFS algorithm handles problems that satisfy the following two assumptions: 1) The cost function $J$ is strictly convex and smooth, which is satisfied by \eqref{eq:opt}. 2)  The nonlinear safety inequality constraints can be written as $x\in \Lambda$ where $\Lambda = \cap_i \Lambda_i$, and $\Lambda_i = \{x : \phi_i(x) \geq 0\}$ where $\phi_i$ is a continuous, piecewise and semi-convex smooth function, which is satisfied by \eqref{eq: safety}. 

 
Given a reference point $x^r$, we compute a convex feasible set $\mathcal{F} := \mathcal{F}(x^r) \subset \Lambda$ around $x^r$.   Note that the convex feasible set with respect to a reference point is not unique. For each constraint $\Lambda_i$, We will find a convex feasible set $\mathcal{F}_i$ and construct the convex feasible set as $\mathcal{F}(x^r) = \cap_i\mathcal{F}_i(x^r)$. The rules of finding $\mathcal{F}_i$ are summarized below: 

\textit{Case 1: $\Lambda_i$ is convex:} 
Define $\mathcal{F}_i = \Lambda_i$. 

\textit{Case 2: The complementary of $\Lambda_i$ is convex:} 
In this case, we can design a convexified $\phi_i(x)$, so that $\phi_i(x) \geq \phi_i(x^r) + \nabla\phi_i(x^r)(x - x^r)$. If $\phi_i$ is not differentiable, we choose $\nabla \phi_i$ as a sub-gradient so that the steepest descent of $J$ in the set $\Lambda$ is always included in the convex set $\mathcal{F}$.  With respect to a reference point $x^r$, the convex feasible set $\mathcal{F}_i$ is defined as 
\begin{equation}\label{eq: cfs case 2}
    \mathcal{F}_i(x^r) = \{ x : \phi_i(x^r) + \nabla\phi_i(x^r)(x - x^r) \geq 0 \}.
\end{equation}


There is a third case considering all other situations, which is not listed here since it is not used in the proposed algorithm. As will be introduced in the experiment section, the safety specification in this paper considers the distance between robot arms and 3-dimensional planes or capsules. Although the complement of the corresponding feasible set in robot configuration space is not always convex, we approximate the convex feasible set using \eqref{eq: cfs case 2} (which might include infeasible points), with the understanding that the approximation error will be minimized when we are approaching the optimal solution. This approach worked successfully in practice and can efficiently find optimal solutions that are strictly feasible, as will be demonstrated in the results section. Nonetheless, we will investigate the feasibility and convergence guarantees of this approximation, as well as consider other sound convexification methods (which do not include infeasible points) in the future. 

The CFS algorithm is guaranteed to converge to local optima if we iteratively minimize the cost function in the convex feasible set and use the intermediate solutions as the reference points to generate the next convex feasible sets. For more details about the convergence and feasibility of the CFS algorithm, the readers are referred to~\cite{liu2018convex}.

\subsection*{Iteractive Convex Optimization for Planning}
\begin{figure}
  \centering
  \includegraphics[scale=0.05]{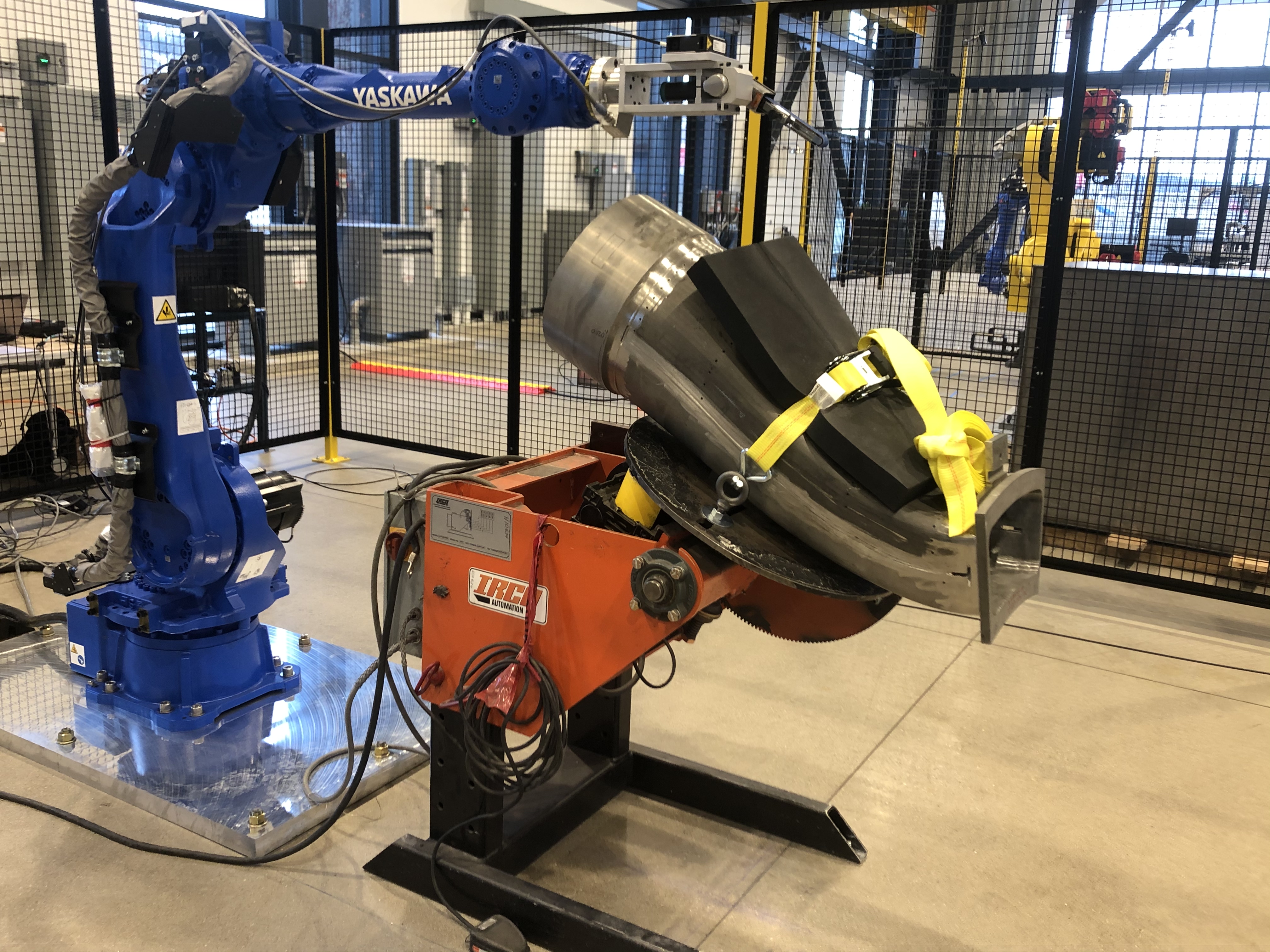}
  \caption{Real weld polishing experiment setup: the blue object is the 6DOF Motoman GP50 robot, while the orange object is the positioner. The silver object is the workpiece to be polished and the weld is inside the workpiece.}
  \label{setup}
\end{figure}
Our goal is to generate contact-rich trajectories in confined environments in real time to safely follow a predefined $m$-dimensional trajectory. We use the CFS algorithm to tackle the workspace inequality constraints and apply \eqref{iela} to linearize the task equality constraints. Then we reduce the original nonlinear and non-convex optimization problem \eqref{eq:opt} to a quadratic program (QP) and solve it iteratively. The proposed motion planning framework, iterative convex optimization for planning (ICOP), is summarized in \cref{alg1}.

The ICOP motion planning algorithm is an incremental method, which solves for a desired trajectory step by step. 

For each outer loop iteration, we get the next desired Cartesian position $C_{next}$, and the configuration $x_{pre}$ from the last step. Next, we use $x_{pre}$ to initialize the temporary reference configuration $x_{ref}$ and the corresponding Cartesian position $C_{ref}$, which will be updated in multiple rounds. Then we enter the SafeTrack inner loop procedure. At each inner loop iteration, we first  compute the corresponding convex feasible set $\mathcal{F}$ in the configuration space. Since the constraint is not convex, according to Case 2 in CFS, we obtain a linear inequality constraint:
\begin{equation}
     \nabla d(x_{ref}, O) x \geq \nabla d(x_{ref}, O) x_{ref} - d(x_{ref},O).
     \label{eq8}
\end{equation}

Secondly, we use \eqref{iela} to linearize the nonlinear contact equality constraint such that the Cartesian position should be $C_{next}$, the resulting equality constraint is: 
\begin{equation}
     Jac(x_{ref},\theta)\cdot x = Jac(x_{ref},\theta)\cdot x_{ref} + C_{next} - C_{ref} - \sigma.
     \label{eq9}
\end{equation}

Finally, we define the optimization objective to minimize the difference of the configuration states between the current reference and last time step. Then we solve the following QP problem with respect to \eqref{eq8} and \eqref{eq9} and update the reference configuration accordingly:
\begin{equation}
\begin{aligned}
& \underset{x}{\text{min}}
& & J(x,x_{ref}) = \| x - x_{ref}\|_Q^2, \\
& \text{s.t.} & &  \nabla d(x_{ref}, O) x \geq \nabla d(x_{ref}, O) x_{ref} - d(x_{ref},O),\\
& & &  Jac(x_{ref},\theta)\cdot x = Jac(x_{ref},\theta)\cdot x_{ref} + C_{next} - C_{ref} - \sigma,\\
& & &  x_{min} \leq x \leq x_{max}.\\
\end{aligned}
\label{eq10}
\end{equation}

 Once the stopping criteria for the SafeTrack procedure are satisfied (line 13 in \cref{alg1}), i.e. the robot at configuration $x_{ref}$ is collision-free and the difference between the the $j$-th robot body point and the desired location is less than a threshold, we add $x_{ref}$ to the planned trajectory $\bm{x}$ and update the configuration from last step $x_{pre}$ to be $x_{ref}$. Note that in practice we set the objective for \eqref{eq10} as $\| x - x_{ref}\|_Q^2$ instead of $\| x - x_{pre}\|_Q^2$, since we observe deadlock issue in terms of optimization result when choosing the latter objective. The underlying reason for causing that deadlock issue is left for future work.

\section*{RESULTS}

\subsection*{Experimental Setup}
This section demonstrates the effectiveness of the proposed algorithm on a weld grinding task using a 6DOF robot manipulator. The application requires contact-rich trajectory generation in confined environments given 1) workspace inequality constraints that the trajectory should be collision free and 2) task equality constraints that the robot end-effector tip should follow the pre-defined welding trajectory. 
\Cref{setup} shows the experiment setup. A YASKAWA Motoman GP50 robot is mounted on the ground. The GP50 robot needs to grind the weld bead inside the workpiece mounted on the positioner. 

\begin{figure}
  \includegraphics[scale=0.5]{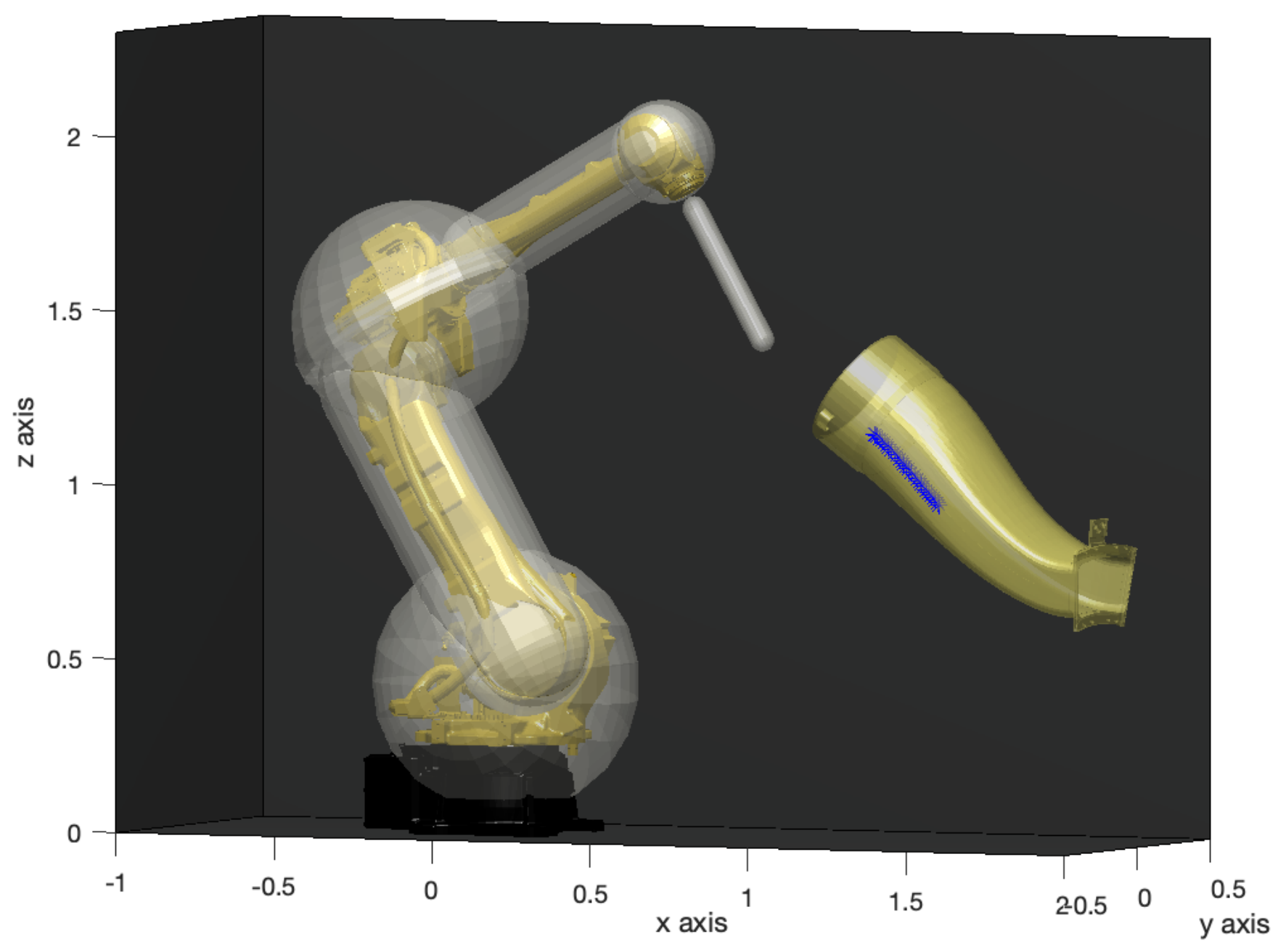}
  \caption{The polishing experiment simulator. The gray transparent capsules are computer aided design models to wrap the robot links. The yellow object on the right is the mounted workpiece. The blue dots inside the workpice are weld points to be polished.}
  \label{sim}
\end{figure}

\begin{figure}
  \centering
  \includegraphics[scale=0.4]{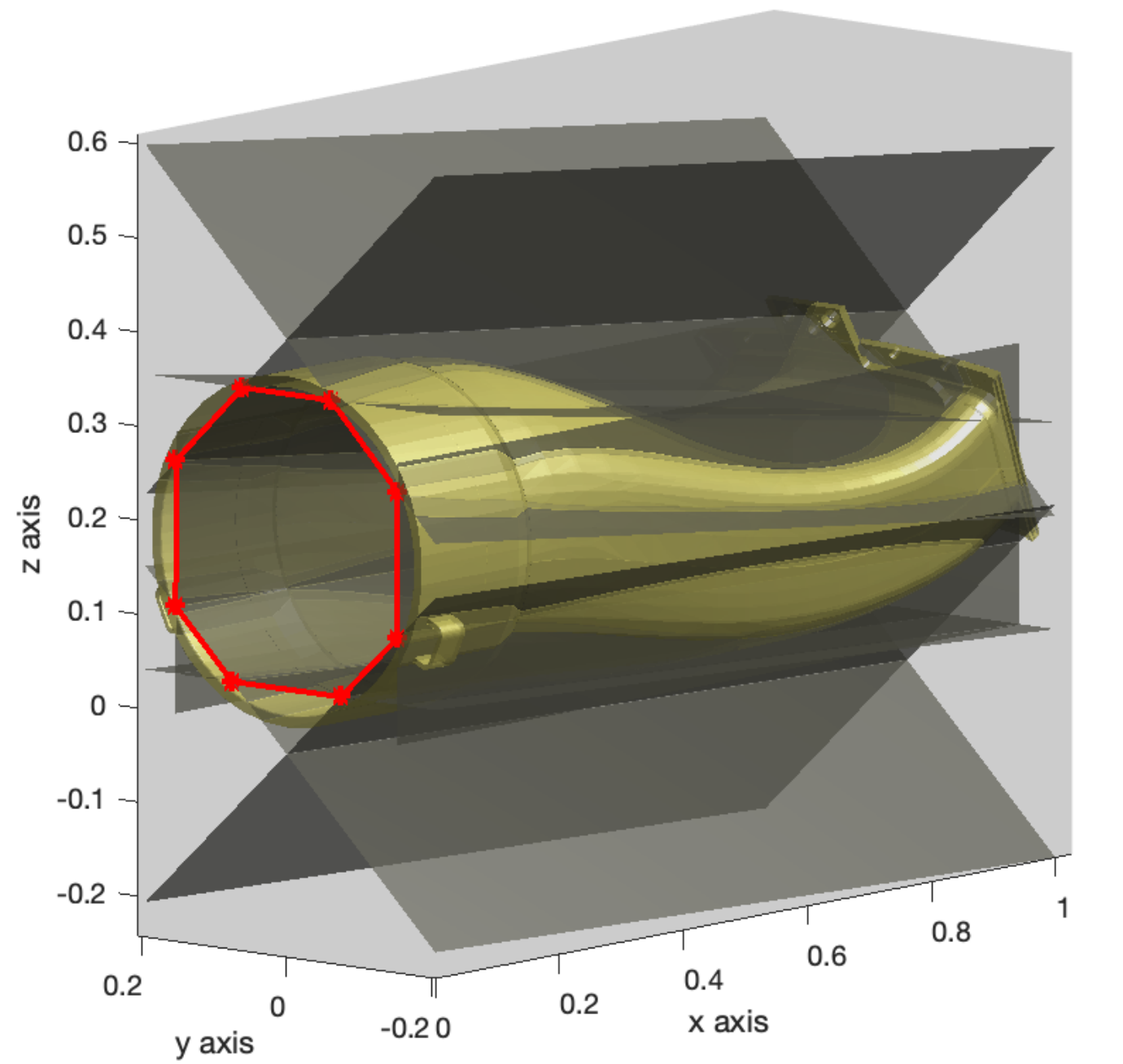}
  \caption{The workpiece obstacle is approximated using $8$ limited-area planes. The red segments denote the intersection lines of different planes on the workpiece entrance surface.}
  \label{planes}
\end{figure}

We first build a simulation environment as shown in \Cref{sim}. This simulator aims to mimic the real world polishing experimental setup. Note that the real world robot links and workpiece have complex shapes. To reduce the computational complexity in the simulation, we use the simplified geometry presentations for both the robot links and the workpiece. In particular, we use six capsules $\Pi_{j=1,2,..,6}$ to wrap the robot links as shown in the left part of \Cref{sim}. To approximate the workpiece, we use $8$ limited-area planes that intersect with each other to construct a polish tunnel that GP50 robot together with the polishing tool should go through. The plane parameters are denoted as $\mathbb{P} \in \mathbb{R}^{4\times 8}$. We need  four parameters to define a single plane. The approximated workpiece is shown in \Cref{planes}. Given the capsules and the limited-area plane representation, we can further define the motion planning constraints as shown below.
 
\subsection*{Task Equality Constraints}

We solve the motion planning problems following the ICOP framework, thus the reference trajectory is generated in an incremental manner. Therefore, at time step $t$, the task equality constraint can be expressed in the following way: the end-effector tip position computed by forward kinematics with respect to the joint configuration $x_t$ should co-locate with the next desired end-effector tip position (weld point) $C_{next}$:
\begin{equation}
\mathbb{FK}(x_t) = C_{next},
\end{equation}
where $\mathbb{FK}(x)$ is the forward kinematics function to get the end-effector tip position in Cartesian space. $C_{next} \in \mathbb{R}^3$  is the desired positions of the weld points in the Cartesian space.

 \begin{table*}[htbp]
 \caption{ Performance comparison among our proposed methods, SQP and interior point in terms of generating collision-free polishing trajectory on four workpiece mounting configurations. $C1,C2,C3,C4$ denote configurations of ($l=1$,$\alpha=0.2\pi$), ($l=135$,$\alpha=0.2\pi$), ($l=15$,$\alpha=0$), and ($l=18$,$\alpha=0.125\pi$), respectively. The average TCP distance between end-effector trajectory and weld path, motion planning computation time, and the average closest distance (safe distance) between robot and obstacle are reported. Bold results highlight the best performance.}
\centering
\begin{tabular}{@{}p{0.04\textwidth}*{9}{L{\dimexpr0.1\textwidth-2\tabcolsep\relax}}@{}}
\toprule
& \multicolumn{3}{c}{TCP distance ($cm$)} &
\multicolumn{3}{c}{Computation time ($s$)} & \multicolumn{3}{c}{Safe distance ($m$)}\\
\cmidrule(r{4pt}){2-4} \cmidrule(l){5-7} \cmidrule(l){8-10}
& proposed & interior point & SQP &  proposed & interior point & SQP & proposed & interior point & SQP\\
\midrule
$C1$ & \bf{0.0031}& 0.0925 & 5.5980 & \bf{12.09$\pm$0.90} & 78.61$\pm$4.51 & 23.88$\pm$2.28 & \bf{0.0350} & 0.0198 & 0.0083\\
$C2$ & \bf{0.0041} & 0.0125 & 3.0945 & \bf{11.92$\pm$1.07} & 6181$\pm$1.16 & 23.79$\pm$0.43 & 0.0202 & \bf{0.0390} & 0.0226\\
$C3$  & \bf{0.0048} & 5.8355 & 9.6341 & \bf{11.30$\pm$1.08} & 32.29$\pm$0.87 & 25.92$\pm$2.40 & \bf{0.0250}  & 0.0026 & 0.0104 \\
$C4$  & \bf{0.0045} & 0.0137 & 3.0955 & \bf{{9.60$\pm$0.37}}  & 47.63$\pm$0.04 & 21.08$\pm$0.58 & \bf{0.0401}  & 0.0354 & 0.0280\\
\bottomrule
\end{tabular}
 \label{summarize}
\end{table*}

\subsection*{Workspace Inequality Constraints}
\begin{figure*}
     \centering
    \begin{subfigure}[t]{0.48\textwidth}
        \raisebox{-\height}{\includegraphics[width=\textwidth]{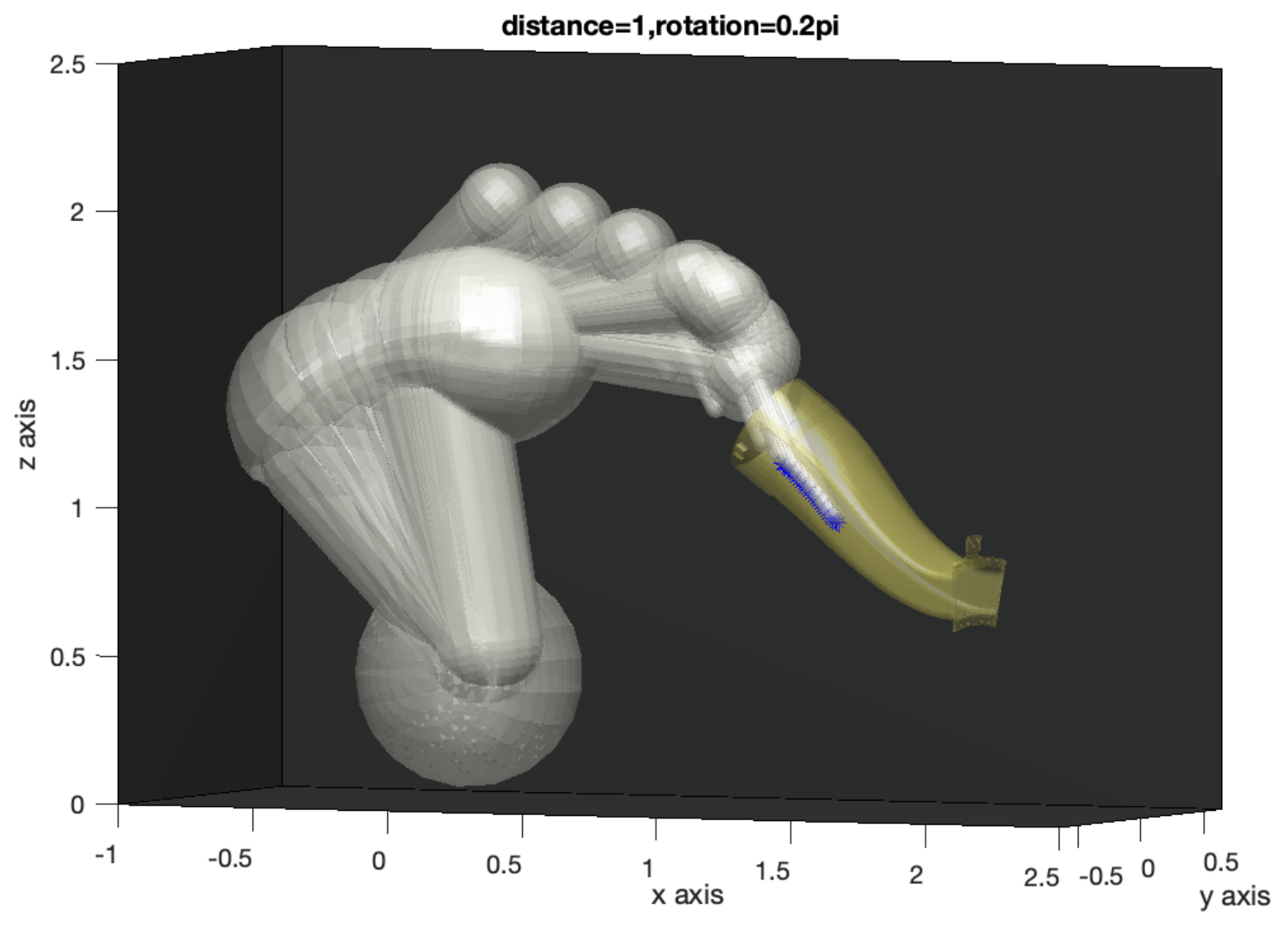}}
        \caption{$l=1$, $\alpha=0.2\pi$}
    \end{subfigure}
    \hfill
    \begin{subfigure}[t]{0.48\textwidth}
        \raisebox{-\height}{\includegraphics[width=\textwidth]{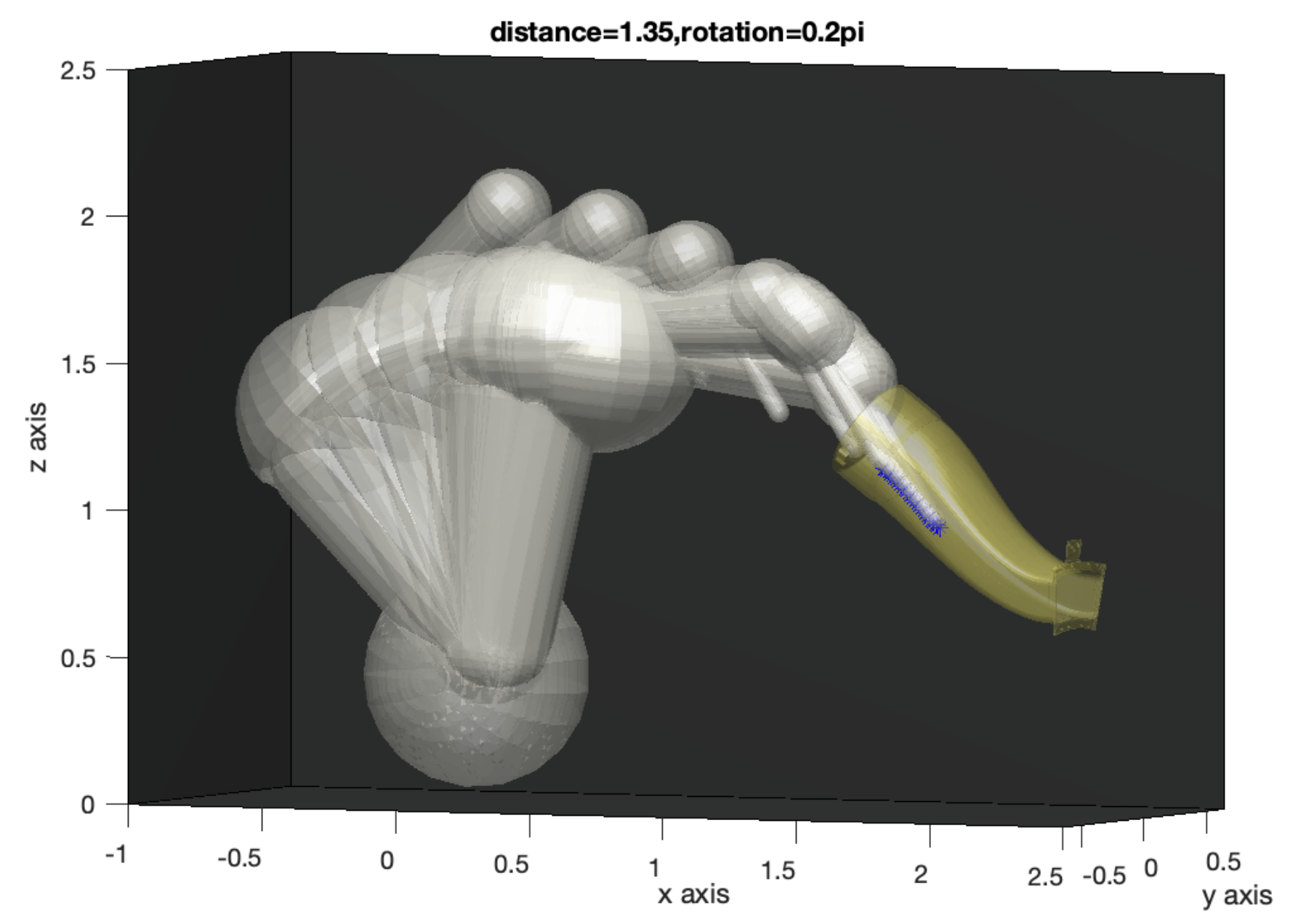}}
        \caption{$l=135$, $\alpha=0.2\pi$}
    \end{subfigure}
    \begin{subfigure}[t]{0.48\textwidth}
        \raisebox{-\height}{\includegraphics[width=\textwidth]{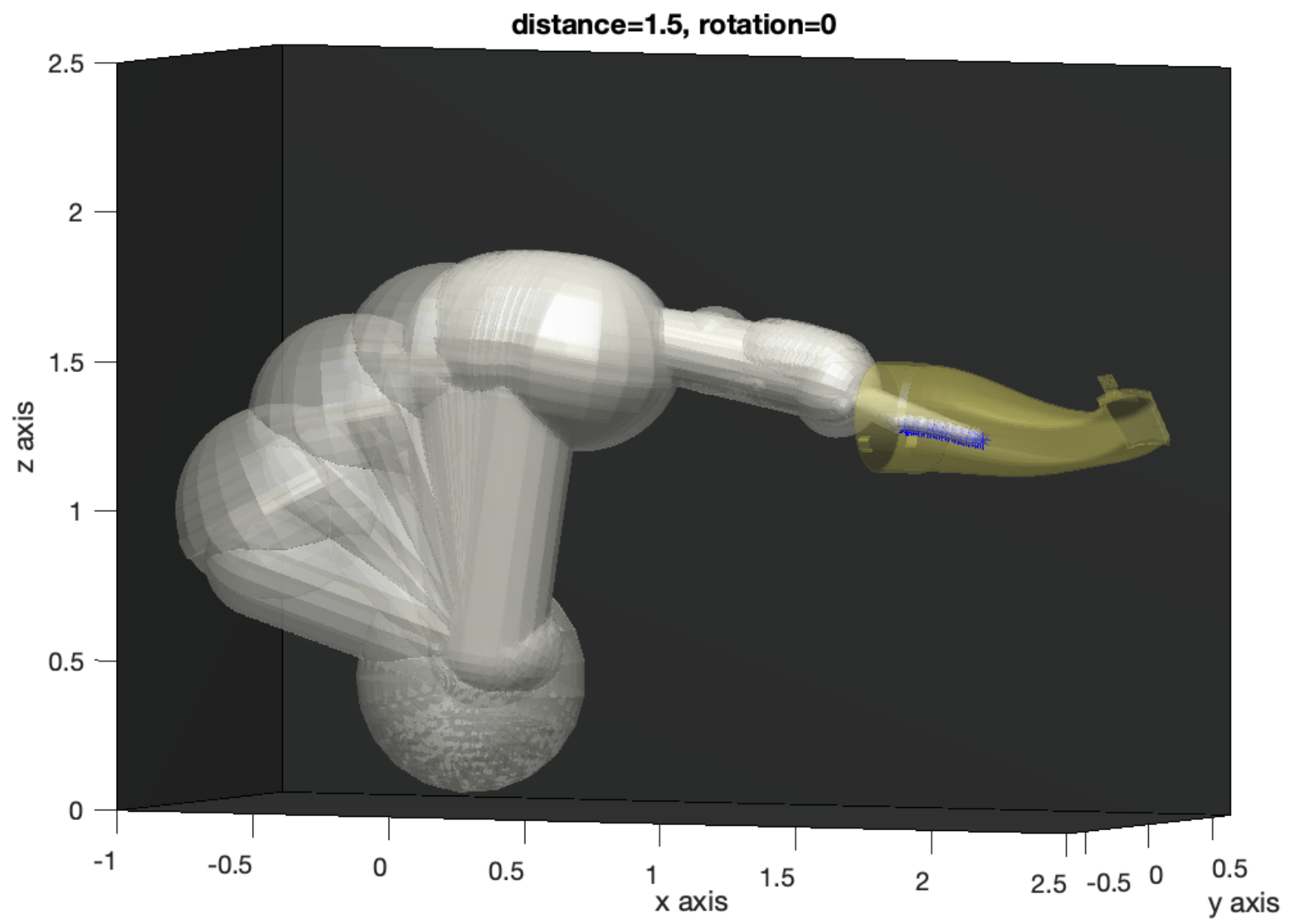}}
    \caption{$l=15$, $\alpha=0$} 
    \end{subfigure}
    \hfill
    \begin{subfigure}[t]{0.48\textwidth}
        \raisebox{-\height}{\includegraphics[width=\textwidth]{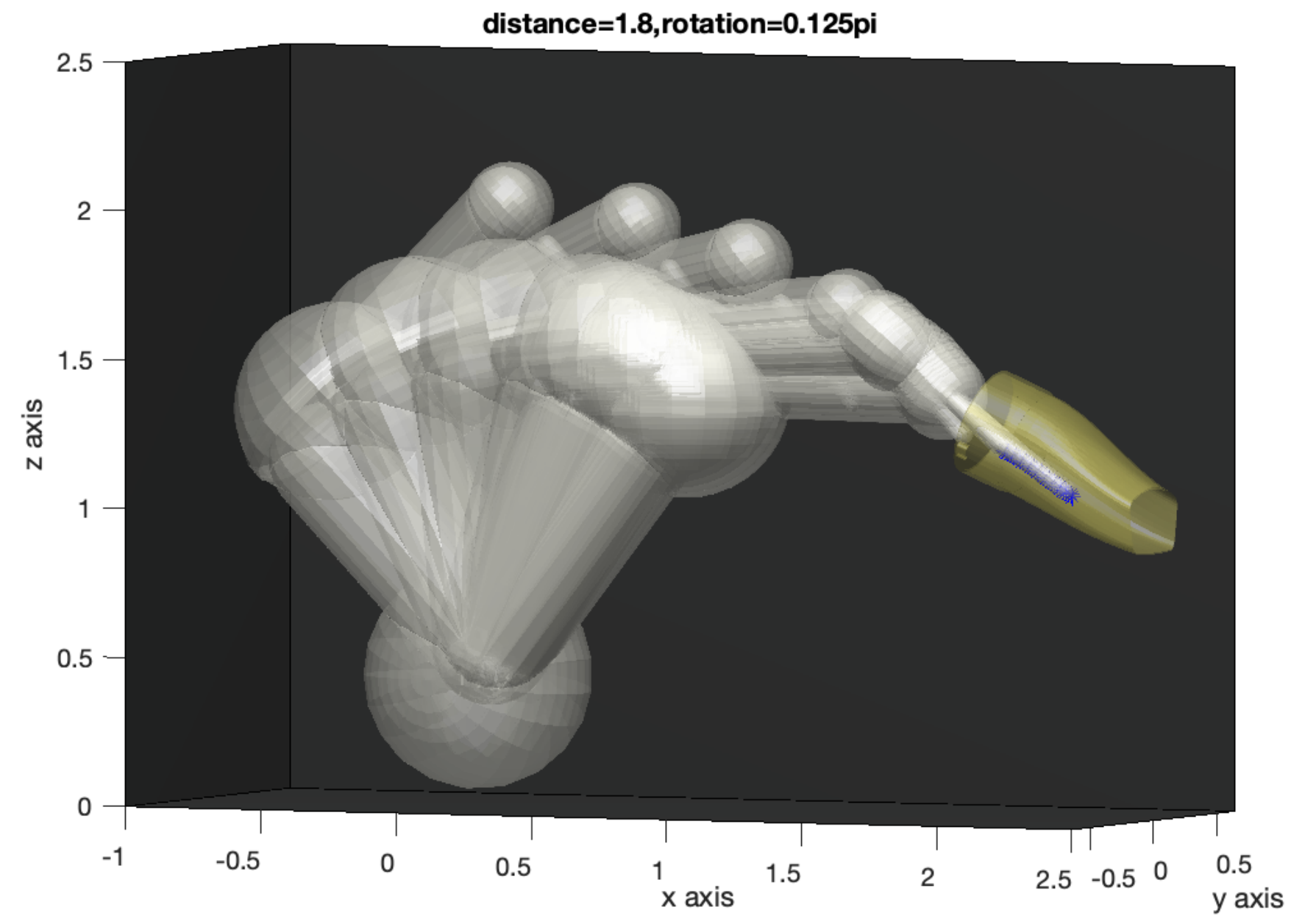}}
    \caption{$l=18$, $\alpha=0.125\pi$} 
    \end{subfigure}
    
     \caption{
     Collision free polishing trajectory generation solution using iterative convex optimization for planning framework on four mounting setting.}
     \label{swipe}
\end{figure*}
To ensure the trajectory is collision-free, at time step $t$, the task inequality constraint can be posed as: the closest distance between workpiece and robot capsules at configuration $x_t$ should be greater than zero. 
\begin{equation}
d(x_t,O) = \min_j(D(\Pi_j^t, O)) > 0,
\end{equation}
where $O$ is the workpiece, $\Pi_j^t$ denotes the capsules at time step $t$, with $j=1,2,...,6$. In the polishing scenario, the safety specification requires that the robot link capsules should not intersect with the limited-area planes. Note that the end-effector tip does not need to follow the safety constraint since it needs to maintain contact with the weld bead. 

For the capsule that does not cross the workpiece entrance surface (\textbf{case 1}), the safety specification requires that the closest distance between the capsule and the tunnel fringe to be greater than zero. As shown in \Cref{planes}, the tunnel fringe can be represented by the collection of segments $\Sigma$, which are intersection lines of different planes on the workpiece entrance surface. Similarly, for the capsule that crosses the workpiece entrance surface (\textbf{case 2}), the safety specification requires the closest distance between the capsule and $\mathbb{P}$ to be greater than zero. Therefore, by replacing obstacle $O$ with $(\mathbb{P},\Sigma)$, we rewrite the distance function as:
\begin{equation}
D(\Pi_j^t,\mathbb{P},\Sigma)= 
\begin{cases}
    \mathcal{D}_1(\Pi_j^t, \Sigma),& \text{if \textbf{case 1}}\\
    \mathcal{D}_2(\Pi_j^t, \mathbb{P}),     & \text{if \textbf{case 2}}
\end{cases},
\end{equation}
where $\mathcal{D}_1(\bullet)$ denotes the function that calculates the closest distance between two line segments. $\mathcal{D}_2(\bullet)$ denotes the function that calculates the smallest distance between line segments and planes. Suppose we define the capsule segments that cross the workpiece entrance surface as working segments. To enforce that robot sticks the polishing tool through the workpiece tunnel, $\mathcal{D}_2(\bullet)$ defines a negative distance value for the working segment that is outside of the workpiece tunnel, and the inequality constraints require the distance value to be strictly positive.

\subsection*{Comparison}

To demonstrate the effectiveness of the proposed method in motion planning given different target and obstacle configurations, we evaluate the proposed algorithm on a collection of four different scenarios. In those scenarios, the workpiece is mounted in different locations relative to the robot. We consider two types of workpiece mounting parameters, translation $l$ along the $x$ axis of the world frame and rotation $\alpha$ about the $y$ axis of the world frame. The four different mountings are illustrated in \Cref{swipe}. 
To verify our algorithm, we compare it with the state-of-the-art nonlinear optimization algorithm interior point and SQP. Both the interior point and SQP algorithm solves the nonlinear equality and inequality constraints using MATLAB \verb+fmincon+ function. The proposed algorithm, interior point and SQP are all implemented using an incremental planning manner. They have the same target and obstacle settings, and the termination conditions and optimization objectives are set to be the same. All the experiments are performed on the MATLAB 2019 platform with a $2.3 GHz$ Intel Core i7 Processor.

We evaluate the performance of task constrained motion planning algorithms using the criteria of the average distance between end-effector tip and desired weld point (TCP distance), the computation time to find the feasible solution, and the average closest distance between the robot and the obstacle (safe distance). The comparison results among interior point, SQP and our method are summarized in \Cref{summarize}. The planning horizon for all the four experiments ($C1,C2,C3$ and $C4$) is $43$, and the equality threshold $\xi$ is set to be 1e-4.

 Compared with interior point and SQP, our method takes the shortest computation time to find the feasible solution, which is generally $4$ to $7$ times faster than interior point and $2$ times faster than SQP. Specifically, for each outer loop iteration, it only takes $3\pm2$ inner loop iteration (SafeTrack procedure in \Cref{alg1}) for our method to find the configuration that satisfies both the equality and inequality constraints, with each inner loop iteration only costs $0.1128\pm0.0893$ seconds. On the other hand, it takes $8\pm2$ inner loop iterations for SQP, and takes $30\pm15$ inner loop iterations for interior point to find a feasible configuration solution. The inner loop iteration time costs are $0.0741\pm0.001$ seconds and $0.0657\pm0.0015$ seconds for SQP and interior point, respectively. Therefore, given the highly nonlinear and non-convex constraints, our proposed iterative convex optimization takes significantly less iterations to find a feasible solution, although the computation cost for each iteration is slightly higher than those of SQP and interior point. It is worth mentioning that our method can find a feasible solution within $3$ seconds when implemented using C++. 

We also observe that our method has the smallest TCP distance across all the four experiments, which demonstrates that our method can better satisfy the task equality constraints. On the contrary, interior point cannot always exactly satisfy the equality constraints or it fails to find the feasible solution. Noteworthy example is in experiment $C3$ where the average TCP distance of interior point is $5.8355 cm$, which means interior point sacrifices the equality constraints satisfaction in order to find the feasible solution. Similarly, we can observe that SQP fails to satisfy the equality constraints across all four experiments. Furthermore, we can observe that our method maintains a better average safe distance cross the majorities of the experiments ($C1, C3$ and $C4$).

 \begin{table}
\caption{Computation time of our proposed methods in terms of different planning horizon with mounting configuration in $C4$.}
 \centering\footnotesize
   \begin{tabular}{p{2.9cm}p{0.5cm}p{0.4cm}p{0.4cm}p{0.6cm}p{0.7cm}p{0.5cm}}
   \hline
   Planning horizon & 14 & 21 & 43 & 82 & 123 & 164\\
   \hline
   Computation time ($s$) & 3.23 & 4.91 & 9.60 & 17.12 & 2189 & 26.84\\
\hline
   \end{tabular}

 \label{horizon}
\end{table}

To demonstrate the scalability of our method in terms of planning horizon and equality threshold, we conduct additional testing using the workpiece mounting configuration in $C4$. We augment the pre-defined end-effector trajectory (planning horizon) using linear interpolation, then down-sample the augmented trajectory to generate cases of different planning horizons. The computation time for different horizons is shown in \cref{horizon}. We observe that the computation time of our proposed method scales linearly with respect to the planning horizon. Next, we fix the planning horizon as $43$ and summarize the computation time for different equality thresholds in \cref{xi}. We observe that the computation time increases as the equality threshold decreases. A smaller equality threshold means higher precision for the end-effector tracking. Thus a good equality threshold should be carefully tuned to meet both the tracking precision and the computation efficiency requirements in practice. 
 
 \begin{table}
\caption{Computation time of our proposed method in terms of different equality threshold $\xi$ with mounting configuration in $C4$ and planning horizon of $43$.}
 \centering
 \footnotesize
   \begin{tabular}{p{3cm}p{0.7cm}p{0.7cm}p{0.7cm}p{0.7cm}p{0.7cm}}
   \hline
   Equality threshold ($m$) & 1e-2 & 1e-3 & 1e-4 & 1e-5 & 1e-6\\
   \hline
   Computation time ($s$) & 3.74 & 4.96 & 9.60 & 15.13 & 18.91\\
\hline
   \end{tabular}

 \label{xi}
\end{table}

The planned results using our method in the four mounting configurations are visualized in \cref{swipe}. We can observe that the planned trajectory is collision-free and the task equality constraints are satisfied. In summary, the proposed method is able to generate trajectories that satisfy the task equality and workspace inequality constraints for the contact-rich trajectory generation problem in confined environments. Moreover, the numerical comparison results demonstrate that, compared with traditional nonlinear optimization methods, our proposed method has the best computational efficiency which satisfies the real time requirements for industry motion planning problems.

\section*{CONCLUSION}
This paper presented an iterative convex optimization motion planning method that can efficiently handle the constraints in motion planning for redundant robotic systems subject to workspace inequality constraints and task equality constraints. The proposed method can generate reference trajectory with high tracking precision and large safety distance in real time. Our planner guarantees the satisfaction of the task constraints and robust planning performance is maintained across different planning scenarios. Planning experiments using YASKAWA Motoman GP50 robot on weld grinding have been presented to demonstrate the effectiveness of the proposed method and the possible application in the industry.
One direction for future work will be to incorporate the adaptive hyper-parameter tuning on the optimization problem for motion planning. Note that the hyper-parameters in the optimization problem is crucial for the success of the optimization-based motion planning algorithm, e.g., the weight matrix in the objective function.
Other extensions of the present approach will be aimed at relaxing the assumption of static obstacle positions. In particular, our ultimate goal is to devise an online version of the present planner under different complex and stochastic environments with time-varying constraints.

\bibliographystyle{asmems4}

\bibliography{asme2e}
\end{document}